\documentclass[10pt,twocolumn,letterpaper]{article}

\usepackage{cvpr}
\usepackage{times}
\usepackage{epsfig}
\usepackage{graphicx}
\usepackage{amsmath}
\usepackage{amssymb}
\usepackage[american]{babel}
\usepackage[ruled,vlined]{algorithm2e}
\usepackage{microtype}

\usepackage[breaklinks=true,letterpaper=true,colorlinks,bookmarks=false]{hyperref}

\cvprfinalcopy 

\def\cvprPaperID{8631} 

\ifcvprfinal\fi

\begin{document}

\title{Episode-based Prototype Generating Network for Zero-Shot Learning}

\author{Yunlong Yu$^{1}$\\
\and
Zhong Ji$^{2}$\\
\and
Jungong Han$^{3}$\\
\and
Zhongfei Zhang$^{4}$\\
\and
$^{1}$College of Information Science \& Electronic Engineering, Zhejiang University, China\\
$^{2}$Tianjin Key BIIT Lab, School of Electrical \& Information Engineering, Tianjin University, China\\
$^{3}$WMG Data Science, University of Warwick, UK\\
$^{4}$Department of Computer Science, Binghamton University, USA
\and
{\tt\small yuyunlong@zju.edu.cn, jizhong@tju.edu.cn, jungong.han@warwick.ac.uk, zhongfei@cs.binghamton.edu}
}

\maketitle
\thispagestyle{empty}

\begin{abstract}
   We introduce a simple yet effective episode-based training framework for zero-shot learning (ZSL), where the learning system requires to recognize unseen classes given only the corresponding class semantics. During training, the model is trained within a collection of episodes, each of which is designed to simulate a zero-shot classification task. Through training multiple episodes, the model progressively accumulates ensemble experiences on predicting the mimetic unseen classes, which will generalize well on the real unseen classes. Based on this training framework, we propose a novel generative model that synthesizes visual prototypes conditioned on the class semantic prototypes. The proposed model aligns the visual-semantic interactions by formulating both the visual prototype generation and the class semantic inference into an adversarial framework paired with a parameter-economic Multi-modal Cross-Entropy Loss to capture the discriminative information. Extensive experiments on four datasets under both traditional ZSL and generalized ZSL tasks show that our model outperforms the state-of-the-art approaches by large margins.
\end{abstract}

\section{Introduction}

\noindent With the renaissance of deep learning, tremendous breakthroughs have been achieved on various visual tasks \cite{he2016deep,fan2019shifting,ren2015faster}. However, the deep learning techniques typically rely on the availability of artificially balanced training data, which poses a significant bottleneck against building comprehensive models for the real visual world. In recent years, Zero-Shot Learning (ZSL)~\cite{lampert2009learning,frome2013devise,xian2016latent,Yu2018Stacked,xie2019attentive} has been attracting a lot of attention due to its potential to address the data scarcity issue.

\begin{figure}[t]
\begin{center}
\includegraphics[width=0.93\columnwidth]{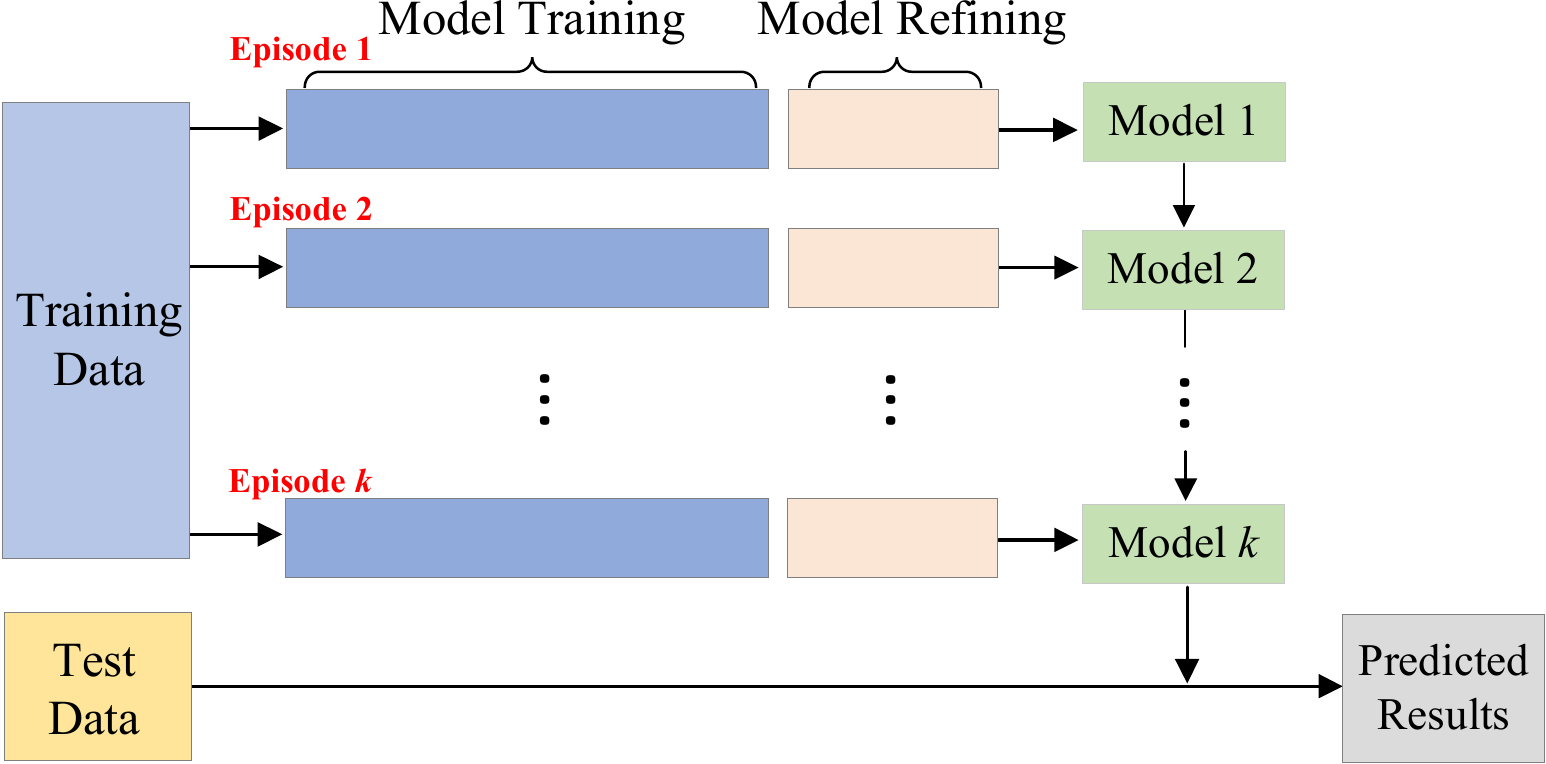}
\end{center}
\caption{An illustration of the episode-based framework for ZSL. The training process consists of a collection of episodes, each of which randomly splits the training data into two class-exclusive subsets: one is used for training the base model, the other one is used for refining the model. The model generalized ability is progressively enhanced as the episodes go on. The test data are predicted with the final model.}
\label{fig1}
\end{figure}

Zero-Shot Learning (ZSL) aims at recognizing unseen classes that have no visual instances during the training stage. Such harsh but realistic scenarios are painful for the traditional classification approaches because there are no labeled visual data to support the parameter training for unseen classes. To tackle this task, the existing methods mostly resort to the transfer learning that assumes the model trained on the seen classes can be applied to the unseen classes, and focus on learning a transferable model with the seen data.

Although promising performances have been achieved, the most existing approaches \cite{akata2013label,akata2015evaluation,xian2016latent,romera2015embarrassingly,frome2013devise,shigeto2015ridge,song2018transductive,sariyildiz2019gradient} dedicated to designing visual-semantic interaction models with the seen classes cannot guarantee to generalize well to the unseen classes, as the seen and unseen classes are located in disjoint domains. Furthermore, the models trained with the seen data favorably guide the unseen test instances to be misclassified into the seen classes, which tends to produce a remarkable imbalanced classification shift issue. The existing generative approaches \cite{kumar2018generalized,li2019leveraging,xian2018feature} transfer the zero-shot classification task to a traditional classification problem via synthesizing some visual features for unseen classes, which can alleviate the above issues to some extent. However, they still struggle in the generalized ZSL task due to the instability in training and mode collapse issues.

Inspired by the success of meta-learning in the few-shot learning task \cite{snell2017prototypical,sung2018learning}, we introduce an episode-based training paradigm to learn a zero-shot classification model for mitigating the above issues. Specifically, the training process consists of a collection of episodes. Each episode randomly splits the training data into two class-exclusive subsets: one support set and one refining set. In this way, each episode mimics a fake zero-shot classification task. The support set is used to train a base model, which builds semantic interactions between the visual and the class semantic modalities. The refining set is used to refine the base model by minimizing the differences between the ground-truth labels and the predicted ones obtained with the base model in a pre-defined space. The model trained in the current episode is initialized with the model parameters learned from the previous episode. As the episodes go on, the base model progressively accumulates ensemble experiences on predicting fake unseen classes, which will generalize well to the real unseen classes. In this way, the gap between the seen and unseen domains can be reduced accordingly. The framework of the whole idea is illustrated in Fig.~\ref{fig1}.

Under the above episode-based training framework, the base model plays an indispensable role in the process of the prediction of unseen classes. In this work, we design an elegant Prototype Generating Network (PGN) as the base model to synthesize class-level visual prototypes conditioned on the class semantic prototypes. As a departure from the existing generative approaches that involve the minimax play games between a generator and a discriminator, our model consists of two generators that map the visual features and the class semantic prototypes into their counterparts and a discriminator that distinguishes between the concatenation of the real visual features and the real class semantic prototypes and the concatenation of the fake counterparts. To capture the discriminative information, we further devise a novel Multi-modal Cross-Entropy Loss to integrate the visual features, class semantic prototypes, and class labels into a classification network. Compared with the existing generative approaches that require an extra assisting classification network with a separate set of learning parameters, our classification network introduces no extra parameters, thus is more efficient.

In summary, our contributions are concluded into the following three-fold.
\begin{enumerate}
  \item To enhance the adaptability of the model, we introduce an episode-based training paradigm for ZSL that trains the models within a collection of episodes, each of which is designed to simulate a fake ZSL task. Through training multiple episodes, the model progressively accumulates a wealth of experiences on predicting the fake unseen classes, which will generalize well to the real unseen classes.

  \item We propose a well-designed prototype generating network to synthesize visual prototypes conditioned on the class semantic prototypes. It aligns the visual-semantic interactions by formulating both the visual prototype generation and the class semantic inference into an adversarial framework and captures the discriminative information with an efficient Multi-modal Cross-Entropy Loss.

  \item Extensive experiments on four benchmarks show that the proposed approach achieves the state-of-the-art performances under both the traditional ZSL and the realistic generalized ZSL tasks.
\end{enumerate}

\begin{figure*}[t]
\begin{center}
\includegraphics[width=14.2cm,height=6.2cm]{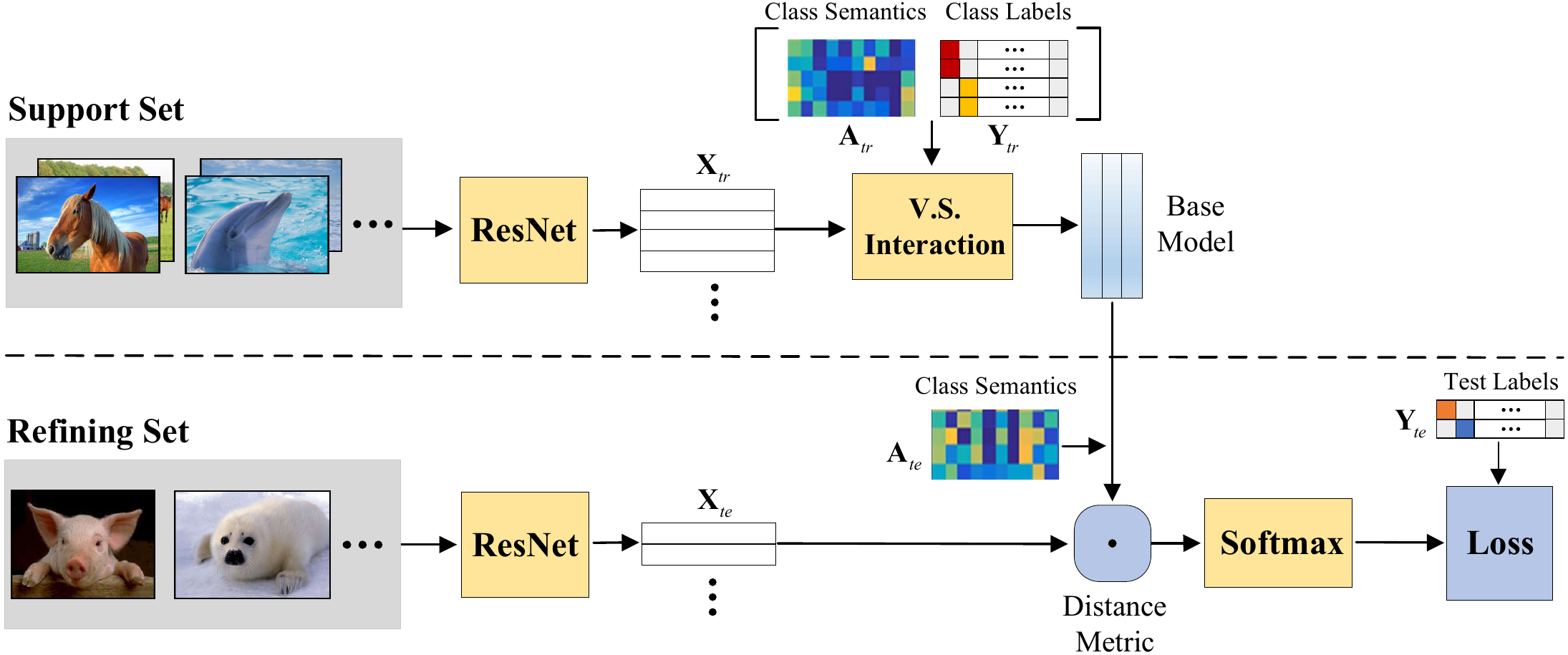}
\end{center}
\caption{Diagram of the proposed approach for one episode training step that consists of a training stage (top) and a refining stage (below). The training stage trains the base model by aligning the visual-semantic interaction (V.S. Interaction). The refining stage first initializes the label prediction of the test data with the trained model in a pre-defined space and then fine-tunes the model by minimizing the differences between the predicted results and the ground-truth labels.}
\label{fig2}
\end{figure*}

\section{Related Work}
In this section, we provide an overview of the most related work on ZSL and episode-based approaches.

\subsection{Generative ZSL}
Recently, the generative approaches predominate in ZSL by exploiting either the existing generative models \cite{goodfellow2014generative,kingma2014auto} or their variations \cite{schonfeld2019generalized,zhu2019learning,atzmon2019adaptive} to synthesize visual features from the class-level semantic features (e.g., attributes and text description embeddings) along with some noises. \cite{xian2018feature,zhu2018generative,li2019leveraging} introduce the Wasserstein generative adversarial network (WGAN) \cite{arjovsky2017wasserstein} paired with a classification network to synthesize visual features for unseen classes such that the ZSL task is transferred to a traditional classification problem. Differently, \cite{zhu2018generative} also introduces a visual pivot regularization to preserve the inter-class discrimination of the generated features while \cite{li2019leveraging} enhances the inter-class discrimination by enforcing the generated visual features to be close to at least one class meta-representations. In contrast to GAN-based approaches, \cite{wang2018zero,schonfeld2019generalized} formulate the feature generation into the variational autoencoder (VAE) \cite{kingma2014auto} model to fit the class-specific latent distribution and highly discriminative feature representations. To combine the strength of VAE and GAN, \cite{xian2019f} develops a conditional generative model to synthesize visual features, which is also extended to exploit the unlabeled instances under the transductive setting via an unconditional discriminator.

Our model is also a generative approach. Instead of synthesizing instance-level visual features, we synthesize class-level visual prototypes conditioned on the class semantic prototypes without extra noise input. Among previous generative approaches, several are closely related to our model.
For example, DEM \cite{zhang2017learning} trains a visual prototype generating network with a three-layer neural network by minimizing the differences between the synthesized visual prototypes and real visual features. In contrast, our approach formulates both the visual prototype generation and class semantic inference into a united framework. Different from \cite{felix2018multi,huang2019generative} that formulate the visual feature generation and class semantic inference in a cycle-consistent manner, our model formulates these two processes with two separable bidirectional mapping networks that are integrated by the discriminator and the classification network, which aligns the visual-semantic interactions better. Furthermore, our approach is trained in an episode-based framework to enhance the adaptability to the unseen classes.

\subsection{Episode-based approach}
Episode-based training strategy has been widely explored in the few-shot learning task \cite{finn2017model,ravi2017optimization,snell2017prototypical,vinyals2016matching} that divides the training process into extensive episodes, each of which mimics a few-shot learning task. However, few researches apply the episode-based training strategy to ZSL.

In this work, we introduce the episode-based paradigm to train the ZSL model. Different from the existing episode-based few-shot approaches, each episode in our approach mimics a zero-shot classification task, which requires to train a base visual-semantic interaction model to achieve the prediction of unseen classes. One related work to ours is RELATION NET \cite{sung2018learning} that also trains a ZSL model in an episode-based paradigm. However, RELATION NET \cite{sung2018learning} learns a general metric space to evaluate the relations between the visual instances and the class semantic features rather than simulating a zero-shot classification task. Another related work is 3ME \cite{felix2019multi} that improves the performance with an ensemble of two different models. Our approach can also be seen as a special ensemble approach that consists of a collection of models. Differently, the models are not equal to vote for the final classification instead of accumulating the previous experiences recursively.

\section{Methodology}
In this section, we first introduce the problem formulation and then report our approach in detail.

\subsection{Problem Formulation}
Suppose that we collect a training sample set
$\mathcal{S} =\{\mathbf{x}_i, \mathbf{a}_i,\mathbf{y}_i\}_{i=1}^N$ that consists of $N$ samples from $M$ seen categories, where $\mathbf{x}_i\in\mathbb{R}^D$ is the $D$-dimensional visual representation (e.g., CNN feature) for the $i$-th instance, $\mathbf{a}_i\in\mathbb{R}^K$ and $\mathbf{y}_i$ are its $K$-dimensional class semantic prototype (e.g., class-level attribute or text description vector) and one-hot class label, respectively. At the test time, in the traditional zero-shot classification setting, the task is to classify a test instance into one of the candidate unseen categories, and in the generalized zero-shot classification setting, the task is to classify the test instance into either a seen or an unseen category.

\subsection{Model}

At the training stage, we introduce an episode-based paradigm for training, which trains the model by simulating multiple zero-shot classification tasks on the seen categories. Each episode matches an individual zero-shot classification task. In each episode, the seen categories $\mathcal{S}$ are randomly split into two class-exclusive sets, one support set $\mathcal{S}^{tr}=\{\mathbf{X}_{tr},\mathbf{A}_{tr},\mathbf{Y}_{tr}\}$ and one refining set $\mathcal{S}^{te}=\{\mathbf{X}_{te},\mathbf{A}_{te},\mathbf{Y}_{te}\}$, where $\mathbf{Y}_{tr}$ and $\mathbf{Y}_{te}$ are disjoint.

As illustrated in Fig.~\ref{fig2}, each episode consists of a training stage and a refining stage. The training stage learns a base model to align the semantic consistency, which is used to predict the unseen classes from the corresponding class semantic prototypes. The refining stage updates the model parameters by minimizing the predicted results and the ground-truth labels. Training each episode can be seen as a process of accumulating experience on zero-shot classification. The experience will be carried forward to the next episode as the episode goes on. After training a collection of episodes, the model is expected to be an expert in predicting unseen classes such that it can generalize well to the real unseen classes. In the following, we introduce the base model and the refining model in an episode in detail.

\begin{figure}[t]
\begin{center}
\includegraphics[width=8.5cm,height=3.4cm]{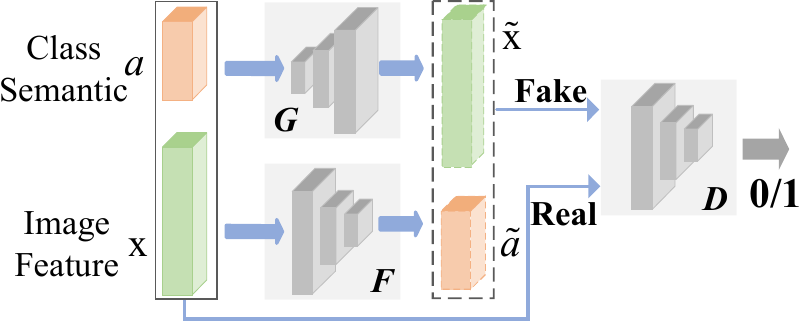}
\end{center}
\caption{The basic model that aligns the semantic consistency across different modalities. The combination of both image feature $\mathbf{x}$ and class semantic prototype $\mathbf{a}$ takes as the real input, while the combination of both the synthesized visual prototype $\mathbf{\tilde{x}}$ and the projected class semantic feature $\mathbf{\tilde{a}}$ as the fake input of the discriminator $D$. Both $F$ and $G$ are mapping networks.}
\label{fig3}
\end{figure}

\subsubsection{Prototype Generating Network}
To address the zero-shot classification task, the learning agent requires learning a base model to infer unseen categories from the corresponding class semantic prototypes. In this paper, we devise a Prototype Generating Network (PGN) to achieve this goal.


For the visual modality, we learn a class semantic inference network $F: \mathbb{R}^D \rightarrow \mathbb{R}^{K}$ to project the image features into the class semantic space by regressing the image features to be close to the corresponding class semantic prototypes, which is formulated as:
\begin{equation}\label{equ1}
 L_{\mathcal{V}\rightarrow \mathcal{A}} = \sum_i \|F(\mathbf{x}_i) - \mathbf{a}_i\|_2^2.
\end{equation}

Similarly, for the class semantic modality, we learn a visual prototype generating network $G: \mathbb{R}^K \rightarrow \mathbb{R}^D$ to project the class semantic prototypes into the visual space. Since each class usually consists of many image instances but corresponds to only one class semantic prototype, the mapping function $G$ can be seen as a one-to-many semantic-to-visual feature generator. The mapping function $G$ is learned by minimizing the distances between the synthesized visual feature $G(\mathbf{a}_i)$ (we call it visual prototype) and the real visual feature $\mathbf{x}_i$.
\begin{equation}\label{equ2}
 L_{\mathcal{A}\rightarrow \mathcal{V}} = \sum_i \|G(\mathbf{a}_i) - \mathbf{x}_i\|_2^2.
\end{equation}

With $F$ and $G$, we can construct the relationships between the visual space and the class semantic space. However, they are independent to each other. To better align the semantic consistency, we introduce the adversarial mechanism to regularize both mapping networks, as illustrated in Fig.~\ref{fig3}. Specifically, we leverage a modified WGAN \cite{gulrajani2017improved} to integrate the projected class semantic vector $\mathbf{\tilde{a}}$ and the real class semantic prototype $\mathbf{a}$ separately to generator and discriminator. The loss is written:
\begin{equation}\label{equ3}
\begin{aligned}
 L_{WGAN}= \mathbb{E}[D(\mathbf{x},\mathbf{a})] - \mathbb{E}[D(\mathbf{\tilde{x}},\mathbf{\tilde{a}})]-\\
 \lambda\mathbb{E}[(\|\nabla_{\mathbf{\hat{x}}}D(\mathbf{\hat{x}},\mathbf{\hat{a}})\|_2-1)^2],
\end{aligned}
\end{equation}
where $\mathbf{\tilde{a}} = F({\mathbf{x}})$ is the inferential class semantic feature; $\mathbf{\tilde{x}} = G(\mathbf{a})$ is the synthetic visual prototype. $\mathbf{\hat{x}}=\tau \mathbf{x}+ (1-\tau) \mathbf{\tilde{x}}$ and $\mathbf{\hat{a}}=\tau \mathbf{a}+ (1-\tau) \mathbf{\tilde{a}}$ with $\tau \thicksim U(0,1)$, $\lambda$ is the penalty coefficient. $D$ denotes the discriminator network. In contrast to the existing GAN-based approaches, the proposed model can be seen as containing two generators and one discriminator, where the generators separately perform on the two different modalities while the discriminator integrates them.

The above model aligns the semantic consistency between the visual features and class semantics. However, training such a model neglects to exploit the discriminative information to distinguish categories, which is essential to the final class prediction. To address this issue, we further propose a multi-modal cross-entropy loss that interweaves the image features, class semantics, and the one-hot class labels into a united framework. With the above model, the class semantic prototypes of all training categories are projected into the visual space to obtain their corresponding class visual prototypes that are pre-stored in a visual feature buffer $G(\mathbf{A}_S)$, where $G(\mathbf{a}_i)$ denotes the class visual prototype of the $i$-th category. The affinities between a visual sample $\mathbf{x}$ and all class visual prototypes could be obtained with their inner products $\mathbf{x}^TG(\mathbf{A}_S)$.  In this way, the probability of the input visual sample $\mathbf{x}$ belonging to the $i$-th category in the visual space can be evaluated with the affinity of visual sample $\mathbf{x}$ matching the $i$-th class semantic vector with the following cross-modal softmax function:
\begin{equation}\label{equ4}
 p_i^\mathcal{V}(\mathbf{x})= \frac{exp(\mathbf{x}^TG(\mathbf{a}_i))}{\sum_{j}exp(\mathbf{x}^TG(\mathbf{a}_j))}.
\end{equation}


Similarly, in the class semantic space, all class semantic vectors are pre-stored in a class semantic buffer $\mathbf{A}_S$ and a visual sample $\mathbf{x}$ is represented as $F(\mathbf{x})$. Therefore, the probability of $\mathbf{x}$ belonging to the $i$-th category in the class semantic space can be defined as
\begin{equation}\label{equ5}
 p_i^\mathcal{S}(\mathbf{x})= \frac{exp(F(\mathbf{x})^T\mathbf{a}_i)}{\sum_{j}exp(F(\mathbf{x})^T\mathbf{a}_j)}.
\end{equation}

Our goal is to maximize the above probabilities in both the visual and class semantic spaces, which can be formulated by minimizing the following Multi-modal Cross-Entropy (MCE) Loss,
\begin{equation}\label{equ6}
 L_{MCE} = -\sum_{\mathbf{x}} \log p_i^\mathcal{V}(\mathbf{x})-\sum_{\mathbf{x}} \log p_i^\mathcal{S}(\mathbf{x}).
\end{equation}

By minimizing Eq.~(\ref{equ6}), the intra-class instances are forced to have higher affinities with their corresponding class semantic prototype than those with the other class semantic prototypes. In this way, the discriminative information can be effectively preserved in both the visual space and class semantic spaces. Compared with the existing generative approaches \cite{li2019leveraging,xian2018feature} that train a softmax classification model with both the real seen visual features and the synthesized unseen visual features, our classification model introduces no extra parameters, which is more efficient and feasible.

Overall, our full objective then becomes,
\begin{equation}\label{equ7}
 \min_G \max_D  L_{WGAN}+ \alpha L_{\mathcal{V}\rightarrow \mathcal{A}}+ \beta L_{\mathcal{A}\rightarrow \mathcal{V}}+\gamma L_{MCE},
\end{equation}
where $\alpha$, $\beta$, and $\gamma$ are hype-parameters to balance each terms.

\subsubsection{Refining Model}

With the trained $G$, the test instance could be classified by searching the nearest generated class visual prototype in the visual space with a pre-defined distance metric. For an unseen instance $\mathbf{x}_t$, its class label is predicted by,
\begin{equation}\label{equ8}
 \hat{y}_t = \arg \min _k (d(\mathbf{x}_t,G(\mathbf{a}_k))),
\end{equation}
where $\mathbf{a}_k$ is the class semantic prototype of the $k$-th unseen class, $G(\mathbf{a}_k)$ is the corresponding generated class visual prototype. $d(\cdot,\cdot)$ denotes a certain distance metric, such as Euclidean or Consine distance.

The base model focuses on building the visual-semantic interactions on the seen classes, which cannot ensure that it generalizes well to the unseen classes in the pre-defined metric space. To enhance the model adaptability to the unseen classes, we refine the part parameters of the base model that are used for predicting unseen classes on the test set $\mathcal{S}^{te}$ in the pre-defined metric space. Specifically, given a distance function $d$, the base model produces a distribution over classes for a test instance $\mathbf{x}_t$ based on a softmax over distance to the class semantic prototypes in the visual space,
\begin{equation}\label{equ9}
 p_G(y=k|\mathbf{x}_t)= \frac{exp(-d(\mathbf{x}_t,G(\mathbf{a}_k)))}{\sum_{k'}exp(-d(\mathbf{x}_t,G(\mathbf{a}_{k'})))},
\end{equation}
where $d(\cdot,\cdot)$ is the distance metric as the same as that in Eq.~(\ref{equ8}). By minimizing the negative log-probability $J(G) = -\log p_G(y=k|\mathbf{x}_t)$ of the true class $k$, the mapping function $G$ is improved for generalizing to the unseen classes in the defined metric space. We observe empirically that the choice of distance metric is vital, as the classification performances with Euclidean distance mostly outperform those with Cosine distance. In the experiments, we report the results with the Euclidean distance, if not specified.

\begin{table}[t]
\begin{center}
\centering
\begin{tabular}{|l@{\hspace{0.1cm}}|c@{\hspace{0.1cm}}|c@{\hspace{0.1cm}}|c@{\hspace{0.1cm}}|c@{\hspace{0.1cm}}|c@{\hspace{0.1cm}}|c@{\hspace{0.1cm}}|}
\hline
Dataset & $\mathcal{K}$ &$\mathcal{Y}_s$ &$\mathcal{Y}_u$ & $\mathcal{X}_a$ & $\mathcal{X}_s$ & $\mathcal{X}_u$ \\
\hline
\hline
AwA1 \cite{lampert2009learning} &85   &40 &10 &30,475 & 5,685 &4,958\\
AwA2 \cite{xian2018zero} &85   &40 &10 &37,322 &5,882 &7,913\\
CUB \cite{wah2011caltech} &1,024 &150 &50 &11,788 &2,967 &1,764\\
FLO \cite{nilsback2008automated} &1,024 &82 &20 &8,189 &5,394 &1,155\\
\hline
\end{tabular}
\end{center}
\caption{The statistics of four benchmark datasets, in terms of class semantic dimensionality $\mathcal{K}$, number of seen classes $\mathcal{Y}_s$, number of unseen classes $\mathcal{Y}_u$, number of all instances $\mathcal{X}_a$, number of test seen instances $\mathcal{X}_s$ and unseen instances $\mathcal{X}_u$. }
\label{tab1}
\end{table}

\begin{table*}
\small
\begin{center}
    \begin{tabular}{|l@{\hspace{0.3cm}}|c@{\hspace{0.28cm}}c@{\hspace{0.28cm}}c@{\hspace{0.28cm}}c@{\hspace{0.28cm}}|c@{\hspace{0.1cm}}c@{\hspace{0.1cm}}
    c@{\hspace{0.1cm}}c@{\hspace{0.1cm}}|c@{\hspace{0.28cm}}c@{\hspace{0.28cm}}c@{\hspace{0.28cm}}c@{\hspace{0.28cm}}|c@{\hspace{0.1cm}}
    c@{\hspace{0.1cm}}c@{\hspace{0.1cm}}c@{\hspace{0.1cm}}|}
    \hline
    \quad &\multicolumn{4}{c|}{AwA1} &\multicolumn{4}{c|}{AwA2} &\multicolumn{4}{c|}{CUB} &\multicolumn{4}{c|}{FLO}\\
    \cline{2-17}
    Method &\textbf{T} &\textbf{u} & \textbf{s} &\textbf{H} &\textbf{T} &\textbf{u} & \textbf{s} &\textbf{H} &\textbf{T} &\textbf{u} & \textbf{s} &\textbf{H } &\textbf{T} &\textbf{u} & \textbf{s} &\textbf{H}\\
    \hline
    \hline
    ALE \cite{akata2013label}   & 59.9 &16.8 &76.1 &27.5  & 62.5 &14.0 &81.8 &23.9   & 54.9 &23.7 &62.8 &34.4  & 48.5  &13.3 &61.6 &21.9\\
    SJE \cite{akata2015evaluation}  & 65.6 &11.3 &74.6 &19.6   & 61.9 &8.0 &73.9 &14.4     & 53.9&23.5 &59.2 &33.6   & 53.4 &13.9 &47.6 &21.5\\
    ESZSL \cite{romera2015embarrassingly} & 58.2 &2.4 &70.1 &4.6   & 58.6 &5.9 &77.8 &11.0  & 53.9 &12.6 &\textbf{63.8} &21.0  & 51.0  &11.4 &56.8 &19.0\\
    DEM \cite{zhang2017learning} &68.4	&32.8	&84.7 &47.3	&67.1 &30.5 &86.4 &45.1 &51.7 &19.6	&57.9 &29.2 &77.8* &57.2* &67.7* &62.0*\\
    GAZSL \cite{zhu2018generative}  & 68.2&29.6 &84.2 &43.8 & 70.2 &35.4 &86.9 &50.3 & 55.8 &31.7 &61.3 &41.8  & 60.5&28.1	&77.4	&41.2\\
    CLSWGAN \cite{xian2018feature} & 68.2& 57.9 &61.4 &59.6   & 65.3 &56.1 &65.5 &60.4   & 57.3 &43.7 &57.7 &49.7   & 67.2   &59.0 &73.9 &65.6\\
    Cycle-UWGAN \cite{felix2018multi} & 66.8&56.9 &64.0 &60.2 & -&- &- &- & 58.6&45.7 &61.0 &52.3  & 70.3&59.2 &72.5 &65.1\\
    SE-ZSL \cite{kumar2018generalized}& 69.5 &56.3 &67.8 &61.5 & 69.2&\textbf{58.3} &68.1 &62.8 & 59.6&41.5 &53.3 &46.7 & - &- &- &-\\
    LisGAN \cite{li2019leveraging}   & 70.6&52.6 &76.3 &62.3 & 70.4*&47.0*&77.6* &58.5* & 58.8&46.5 &57.9 &51.6 & 69.6  &57.7 &83.8 &68.3\\
    f-VAEGAN-D2 \cite{xian2019f}  & 71.1&57.6 &70.6 &63.5 & -&- &- &-  & 61.0&48.4 &60.1 &53.6  & 67.7&56.8 &74.9 &64.6\\
    CADA-VAE \cite{schonfeld2019generalized} &62.3 &57.3 &72.8 &64.1 &64.0 &55.8 &75.0 &63.9 &60.4 &51.6 & 53.5 &52.4 & -&-&-&-\\
    ABP \cite{zhu2019learning} &69.3 &57.3 &67.1 &61.8 &70.4 &55.3 &72.6 &62.6 &58.5 &47.0 &54.8 &50.6 &- &- &- &-\\
    RELATION NET \cite{sung2018learning} & 68.2 &31.4 &\textbf{91.3} &46.7 & 64.2 &30.0 &\textbf{93.4} &45.3 & 55.6&38.1 &61.1 &47.0  & 78.5* &50.8* &\textbf{88.5}* &64.5*\\
    3ME \cite{felix2019multi} &65.6 &55.5 &65.7& 60.2 & - &- &- &- & 71.1 & 49.6 & 60.1 & 54.3 & 83.9 & 57.8 & 79.2 & 66.8\\
    \hline
    \textbf{E-PGN} (Ours) & \textbf{74.4}&\textbf{62.1} &83.4 &\textbf{71.2}& \textbf{73.4}& 52.6 &83.5 &\textbf{64.6} & \textbf{72.4} & \textbf{52.0} &61.1 &\textbf{56.2}  & \textbf{85.7} &\textbf{71.5} &82.2 &\textbf{76.5}\\
    \hline
    \end{tabular}
    \end{center}
    \caption{\upshape Performance (in \%) comparisons for both traditional and generalized ZSL in terms of average per-class top-1 accuracy (\textbf{T}), unseen accuracy (\textbf{u}), seen accuracy (\textbf{s}), and their harmonic mean (\textbf{H}). $^*$ indicates the results obtained by ourselves with the codes released by the authors. The best results are marked in boldface.}
    \label{tab2}
\end{table*}

The model PGN trained with episode-based framework is short for E-PGN. The training process of E-PGN is summarized in Algorithm~\ref{alg1}
\begin{algorithm}
\label{alg1}
\caption{Proposed E-PGN approach.}
\KwIn{The seen category set $\mathcal{S}$, the hyper-parameters $\alpha$, $\beta$, and $\gamma$.}
\KwOut{Visual prototype generating network $G$.}
Initialize the parameters of both $F$ and $G$. \\
\While{not done}
{
Randomly sample $\mathcal{S}^{tr}$ and $\mathcal{S}^{te}$ from $\mathcal{S}$\;
\For{samples in $\mathcal{S}^{tr}$}
{
Optimize $F$ and $G$ by Eq.~(\ref{equ7})\;
}
\For{samples in $\mathcal{S}^{te}$}
{
Calculate probability distribution by Eq.~(\ref{equ9})\;
Update $G$ by minimizing the negative log-probability.
}
}
return The parameters of $G$.
\end{algorithm}

\section{Experiments}
In this section, we conduct experiments to evaluate the effectiveness of the proposed model. We first document the datasets and experimental settings and then compare E-PGN with the state-of-the-art. Finally, we study the properties of the proposed E-PGN with a serious of ablation experiments.

\subsection{Datasets and Experimental settings}

\textbf{Datasets.} Among the most widely used datasets for zero-shot classification, we select two coarse-grained datasets, namely Animals with Attributes (AwA1) \cite{lampert2009learning}, Animals with Attributes2 (AwA2) \cite{xian2018zero}, and two fine-grained datasets, i.e., Caltech-UCSD Birds-200-2011 (CUB) \cite{wah2011caltech} and Oxford Flowers (FLO) \cite{nilsback2008automated}. AwA1 and AwA2 consist of different visual images from the same 50 animal classes, each class is annotated with 85-dimensional semantic attributes. CUB and FLO respectively contain 200 bird species and 102 flower categories. As for the class semantic representations of both CUB and FLO datasets, we average the 1,024-dimensional character-based CNN-RNN \cite{reed2016learning} features extracted from the fine-grained visual descriptions (10 sentences per image). We adopt the standard zero-shot splits provided by \cite{xian2018zero} for AwA1, AwA2, and CUB datasets. For FLO dataset, we use the splits provided by \cite{nilsback2008automated}. A dataset summary is given in Table~\ref{tab1}.

\textbf{Evaluation Protocol.} In this work, we evaluate our approach on both traditional ZSL and generalized ZSL tasks. For the traditional ZSL task, we apply the extensively used average per-class top-1 accuracy (\textbf{T}) as the evaluation protocol. For the generalized ZSL task, we follow the protocol proposed in \cite{xian2018zero} to evaluate the approaches with both seen class accuracy \textbf{s} and unseen class accuracy \textbf{u}, as well as their harmonic mean \textbf{H}.

\textbf{Implementation settings.} Following \cite{xian2018zero,xian2018feature}, we use the top pooling units of the ResNet-101 \cite{he2016deep} pre-trained on ImageNet-1K as the image features. Thus, each input image is represented as a 2,048-dimensional vector. As a pre-processing step, we normalize the visual features into [0,~1]. In terms of the model architecture, we implement $F$, $G$, and $D$ as simple three-layer neural networks with 1,800, 1,800, and 1,600 hidden units. Both $F$ and $D$ apply ReLU as the activation function on both the hidden layer and the output layer, both of which follow a dropout layer. While developing the model, we have observed that by applying the tanh activation function for the hidden layer of $G$ would obtain more stable and better results. In terms of the learning rate of the base model, we set $5e^{-5}$ for AwA1, CUB, and FLO datasets, and $2e^{-4}$ for AwA2 dataset. For all datasets, we set the learning rate of the refining model as $1/10$ of the original base model. In each episode, the base model is trained for 100 epochs by stochastic gradient decent using the Adam optimizer and a batch size of 128 for AwA1 dataset and 32 for the other datasets. The refining model in each episode is trained for 10 epochs using the same optimizer and batch size as the base model. Our model is implemented using TensorFlow framework. The code is available at \footnote {\url{https://github.com/yunlongyu/EPGN}}.

\begin{table*}[t]
\begin{center}
    \centering
    \begin{tabular}{|l@{\hspace{0.2cm}}|c@{\hspace{0.25cm}}c@{\hspace{0.25cm}}c@{\hspace{0.25cm}}c@{\hspace{0.25cm}}|c@{\hspace{0.25cm}}c@{\hspace{0.25cm}}
    c@{\hspace{0.25cm}}c@{\hspace{0.25cm}}|c@{\hspace{0.25cm}}c@{\hspace{0.25cm}}c@{\hspace{0.25cm}}c@{\hspace{0.25cm}}|c@{\hspace{0.25cm}}
    c@{\hspace{0.25cm}}c@{\hspace{0.25cm}}c@{\hspace{0.25cm}}|}
    \hline
    \quad &\multicolumn{4}{c|}{AwA1} &\multicolumn{4}{c|}{AwA2} &\multicolumn{4}{c|}{CUB} &\multicolumn{4}{c|}{FLO}\\
    \cline{2-17}
    Method&\textbf{T} &\textbf{u} & \textbf{s} &\textbf{H} &\textbf{T} &\textbf{u} & \textbf{s} &\textbf{H} &\textbf{T} &\textbf{u} & \textbf{s} &\textbf{H } &\textbf{T} &\textbf{u} & \textbf{s} &\textbf{H}\\
    \hline
    \hline
    \textbf{PGN} & 72.2 &52.6&\textbf{86.3} &65.3   &71.2&48.0 &83.6 &61.0  & 68.3 &48.5 &57.2 &52.5  & 81.4  &63.6 &77.8 &70.0\\
    \textbf{E-PGN} (5)   & 72.2 &57.2 &83.8 &68.0    & 73.5 &51.2 &83.0 &63.3      & 70.4 &50.5 &59.0 &54.4  & 84.2  &67.7 &79.6 &73.2\\
    \textbf{E-PGN} (10) & \textbf{74.4}&62.1 &83.4 &\textbf{71.2}& 73.4& \textbf{52.6} &83.5 &\textbf{64.6}  & \textbf{72.4} & \textbf{52.0} &\textbf{61.1} &\textbf{56.2}  &\textbf{85.7} &\textbf{71.5} &\textbf{82.2} &\textbf{76.5}\\
    \textbf{E-PGN} (15) & 73.8&\textbf{62.2} &82.9 &71.1& \textbf{74.2}& 50.5 &\textbf{84.1} &63.1 & 69.6 & 51.5 &57.4 &54.3  & 85.3 &70.5 &80.4 &75.2\\
    \hline
    \end{tabular}
    \end{center}
    \caption{\upshape Performance (in \%) comparisons of the number of the selected mimetic unseen classes in each episode. PGN indicates the model trained without episode-based paradigm.}
    \label{tab3}
\end{table*}
\subsection{Comparing State-of-The-Art Approaches}
Table~\ref{tab2} describes the classification performances of E-PGN and fourteen competitors including three discriminative approaches \cite{akata2013label,akata2015evaluation,romera2015embarrassingly}, nine generative approaches \cite{zhang2017learning,zhu2018generative,xian2018feature,felix2018multi,kumar2018generalized,li2019leveraging,xian2019f,schonfeld2019generalized,zhu2019learning}, one episode-based approach \cite{sung2018learning}, and one ensemble approach \cite{felix2019multi}.

From Table~\ref{tab2}, we observe that the proposed E-PGN achieves significant improvements over the state-of-the-art in terms of both \textbf{T} and \textbf{H} on four datasets. Specifically in \textbf{T} metric, the overall accuracy improvement on AwA1 increases from 71.1\% to 74.4\%, on AwA2 from 70.4\% to 73.4\%, on CUB from 71.1\% to 72.4\%, and on FLO from 83.9\% to 85.7\%, i.e., all quite significant. Remarkably, E-PGN achieves 71.2\% and 76.5\% for \textbf{H} metric on AwA1 and FLO datasets, which marginally improves the second-best performance by 7.1\% and 8.2\%. On AwA2 and CUB datasets, the proposed E-PGN also gains improvements from 63.9\% to 64.6\% and from 54.3\% to 56.2\%, respectively. Compared with the other episode-based approach RELATION NET \cite{sung2018learning}, our E-PGN achieves significant improvements on four datasets, which indicates that our mimetic strategy captures more discriminative transfer knowledge than learning the distance metric strategy. Compared with the other ensemble approach 3ME  \cite{felix2019multi}, our E-PGN also has obvious improvements under different metrics across different datasets.

We also observe that the seen classification accuracy \textbf{s} is much better than unseen classification accuracy \textbf{u}, which indicates that the unseen test instances tend to be misclassified into the seen classes. This classification shifting issue is ubiquitous across all the existing approaches. From the results, we observe that the generative approaches alleviate this shift issue to some extent, resulting in the improvement of \textbf{H} measure. However, those approaches balance the differences between the seen class accuracy and the unseen class accuracy via decreasing the seen class accuracy while improving the unseen class accuracy, which is not desirable in practice. In contrast, our E-PGN is more robust than the competitors, which substantially boosts the harmonic mean \textbf{H} via improving the unseen class accuracy while maintaining the seen class accuracy at a high level. Our performance improvement is benefited from the progressive episode-training strategy paired with the effective base model.

\subsection{Further Analysis}
\subsubsection{Impact of episode-based paradigm} In the first experiment, we evaluate the impact of the episode-training scheme and how the number of selected mimetic unseen classes in each episode affects the performances on different datasets. To do so, we vary the number of selected mimetic unseen classes from 0 to 15 in intervals of 5. It should be noted that the case where the number of selected mimetic unseen classes equaling 0 indicates the approach trained without the episode-based paradigm and the optimization process degenerates to the traditional batch-based training strategy.
\begin{figure}[t]
\centering
\includegraphics[width=8.1cm,height=6.5cm]{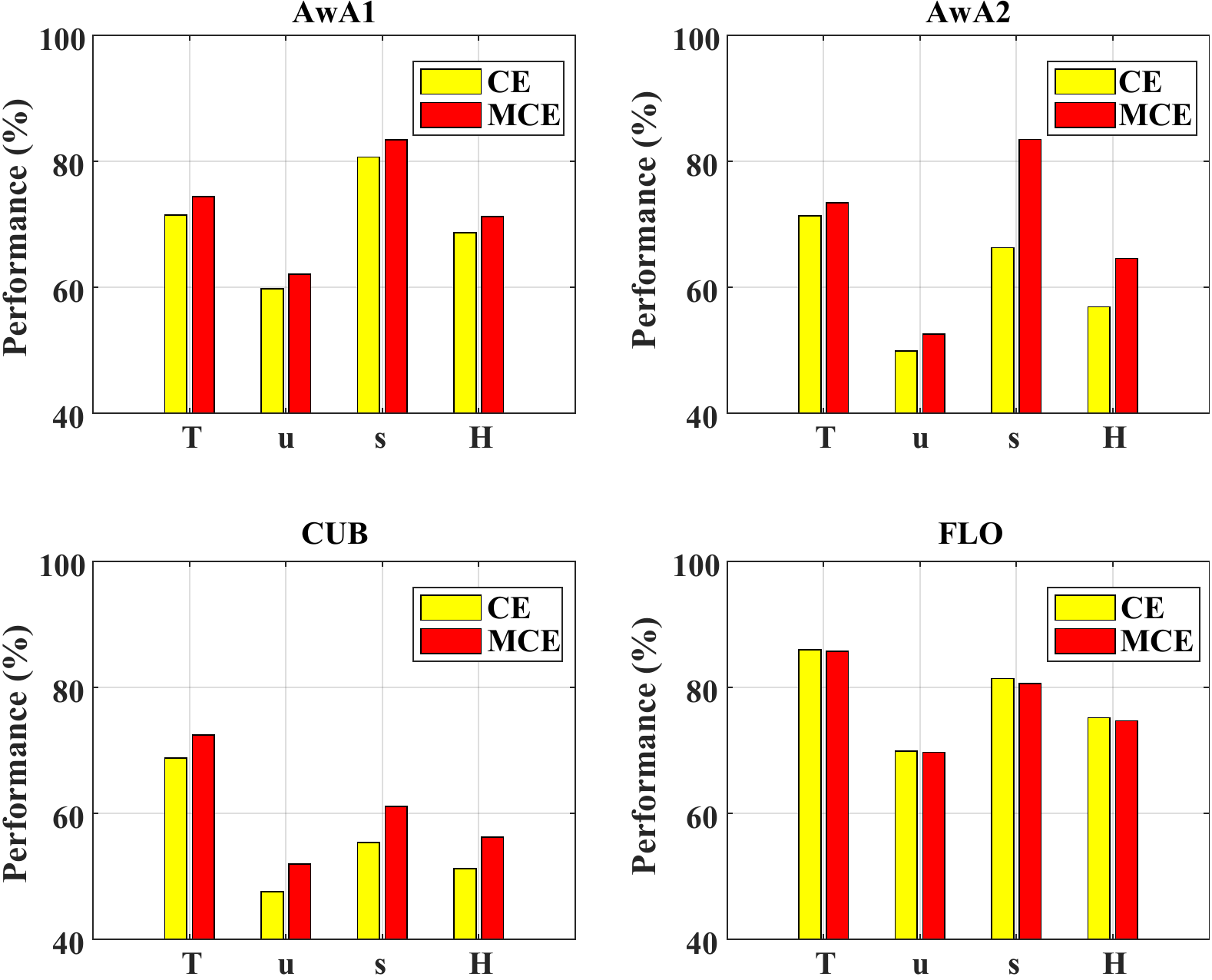}
\caption{Traditional and generalized zero-shot classification results with traditional cross-entropy loss (short for CE) and multi-modal cross-entropy loss (short for MCE) on four datasets.}
\label{fig4}
\end{figure}

According to the results in Table~\ref{tab3}, we observe that E-PGN mostly performs better than PGN on four datasets under different metrics except \textbf{s} on AwA1 dataset, which indicates the effectiveness of the proposed episode-based training strategy. Compared with PGN, the E-PGN may spoil the whole training structure to some extent, but can progressively accumulate the knowledge on how to adapt to novel classes with the episode-based training paradigm, and thus better results are obtained. Besides, we also observe that the number of the selected mimetic unseen classes greatly impacts the classification performances. Specifically, E-PGN~(10) basically beats E-PGN~(5) on four datasets. However, with the further increase of the number, the performances tend to decrease, we speculate that the reason is that when more mimetic unseen classes are selected for refining, fewer training classes are left for training the base model, leading to unsatisfied initialization for the prediction of the mimetic unseen classes.
\begin{table*}[t]
\begin{center}
\begin{tabular}{|ccc| c@{\hspace{0.27cm}}c@{\hspace{0.27cm}}c@{\hspace{0.27cm}}c@{\hspace{0.27cm}}|  c@{\hspace{0.27cm}}c@{\hspace{0.27cm}}c@{\hspace{0.27cm}}c@{\hspace{0.27cm}}|  c@{\hspace{0.27cm}}c@{\hspace{0.27cm}}c@{\hspace{0.27cm}}c@{\hspace{0.27cm}}|  c@{\hspace{0.27cm}}c@{\hspace{0.27cm}}c@{\hspace{0.27cm}}c@{\hspace{0.27cm}}|}
\hline
\multicolumn{3}{|c|}{\quad}& \multicolumn{4}{c|}{AwA1} & \multicolumn{4}{c|}{AwA2} & \multicolumn{4}{c|}{CUB} &\multicolumn{4}{c|}{FLO}\\
\cline{4-19}
$\alpha$   &$\beta$  &$\gamma$ &\textbf{T} &\textbf{u} & \textbf{s} &\textbf{H} &\textbf{T} &\textbf{u} & \textbf{s} &\textbf{H} &\textbf{T} &\textbf{u} & \textbf{s} &\textbf{H } &\textbf{T} &\textbf{u} & \textbf{s} &\textbf{H} \\
\hline
\hline
\quad &\checkmark &\checkmark & 73.1 &60.3&82.3 &69.6  & 72.6 &51.3 &81.6 &63.0   & 71.2 &50.9 &59.1 &54.7  & 85.0  &69.2 &79.7 &74.1\\
\checkmark &\quad &\checkmark & 73.8 &61.0 &83.1 &70.4 & 72.2 &52.5 &82.7 &64.3  & 67.2 &45.9 &55.9 &50.4  & 85.8  &69.4 &82.0 &75.2\\
\checkmark &\checkmark &\quad & 70.8&56.2 &82.2 &66.8& 70.9& 43.2 &79.9 &56.1 & 66.8 & 45.2 &52.5 &48.8  & \textbf{86.2} &70.0 &79.2 &74.4\\
\quad &\quad &\checkmark & 72.1 &56.2&81.5 &66.5  & 71.2 &48.5 &\textbf{84.0} &61.5   & 70.3 &50.0 &57.5 &53.5  & 85.6 &71.3 &80.5 &75.6\\
\checkmark &\checkmark &\checkmark  &\textbf{74.4}&\textbf{62.1} &\textbf{83.4} &\textbf{71.2} & \textbf{73.4}& \textbf{52.6} &83.5 &\textbf{64.6} & \textbf{72.4} & \textbf{52.0} &\textbf{61.1} &\textbf{56.2}  & 85.7 &\textbf{71.5} &\textbf{82.2} &\textbf{76.5}\\
\hline
\end{tabular}
\end{center}
\caption{Ablation study of the E-PGN components on four datasets. The best results are marked in boldface. }
\label{tab4}
\end{table*}

\subsubsection{Performance impacts of E-PGN components}
In this study, we quantify the benefits of the different components in E-PGN on the performances. In the proposed E-PGN model, except for the base adversarial loss, there are three components: two regression losses and one multi-modal classification loss. Each loss is controlled by a hyper-parameter, i.e., $\alpha$, $\beta$, and $\gamma$. We select the values of the hyper-parameters only from 0 and 1. When the value of a hyper-parameter equals 1, its corresponding component is ``switch on", otherwise is ``switch off". The performance differences between the two scenarios reveal the effects of the component.

From the results illustrated in Table~\ref{tab4}, we observe that the model with all three components mostly achieves the best performances for fourteen out of sixteen metrics, which indicates that the three calibration terms complement each other. Besides, we observe that the performances of the model without MCE loss ($\gamma=0$) degrade significantly on three out of four datasets, which reveals that the MCE loss contributes significantly to the classification performance. 



\subsubsection{Impact of classification network}
To further validate the superiority of the proposed multi-modal cross-entropy loss, we compare our E-PGN against the method with the traditional cross-entropy loss. From the results illustrated in Fig.~\ref{fig4}, we observe that the proposed E-PGN with MCE loss performs much better than the counterpart with traditional Cross-Entropy (CE) loss on AwA1, AwA2, and CUB datasets, and performs neck to neck on FLO dataset. We argue that the superiority is due to that the MCE loss encodes with the class semantic information into the classification module, which both preserves the discriminative information and enhances the visual-semantic consistence. Besides, compared with the model with the traditional CE loss, the model with the MCE loss introduces no extra training parameters, which is more efficient.
\begin{figure}[t]
\centering
\includegraphics[width=8.1cm,height=6.3cm]{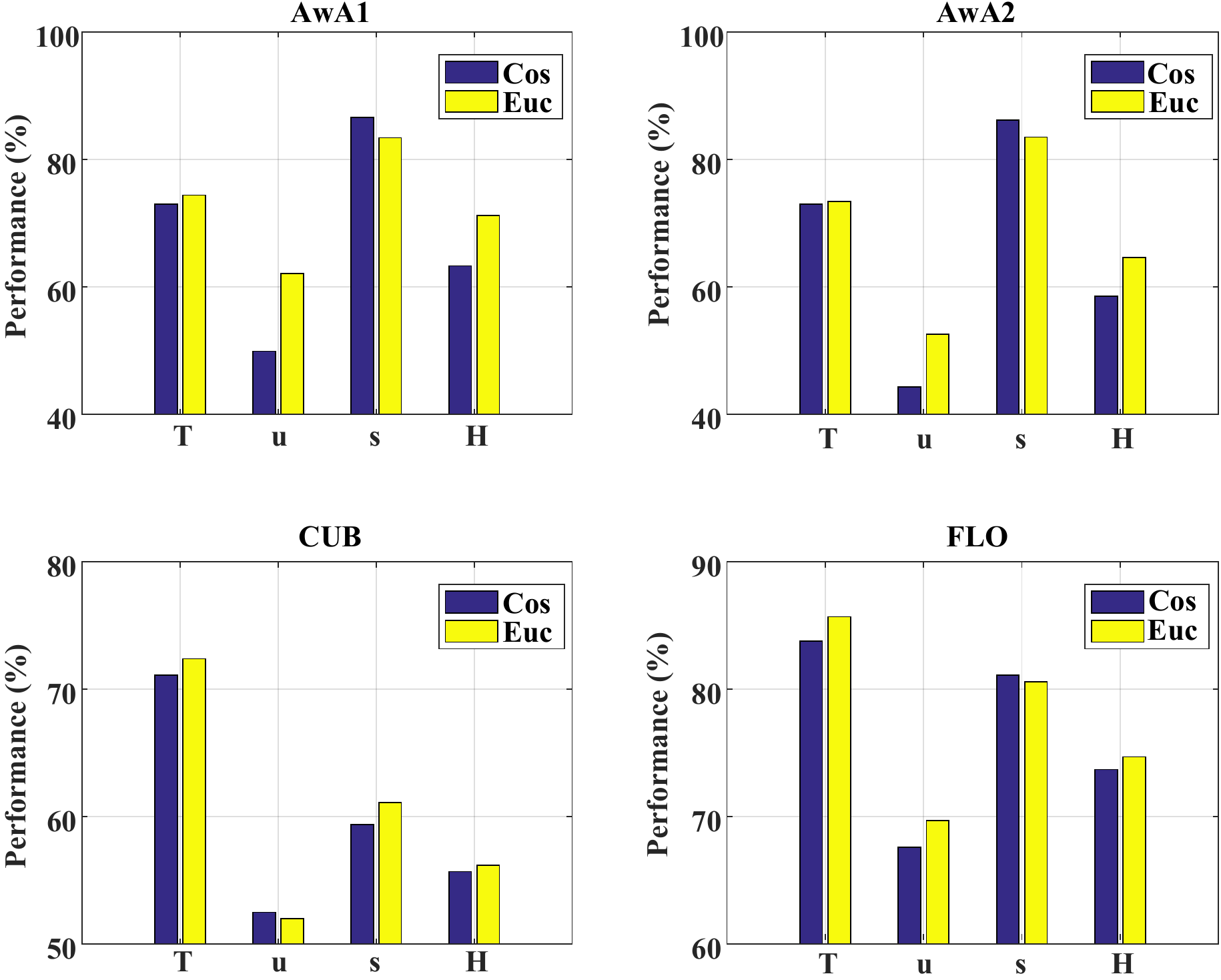}
\caption{Traditional and generalized zero-shot classification results with Euclidean distance (short for Euc) and Cosine distance (short for Cos) on four datasets.}
\label{fig5}
\end{figure}

\subsubsection{Impact of distance metric} In this experiment, we investigate how the distance metric affects the classification performance. In Fig.~\ref{fig5}, we compare Cosine vs. Euclidean distance under different metrics on four datasets. We observe that the performances obtained in the Euclidean space are significantly better than those obtained in the Cosine space under most cases, indicating that the Euclidean distance is more suitable to our approach. The inferior performances obtained in Cosine space may be due to that the Cosine distance is not a Bregman divergence~\cite{snell2017prototypical}.

\section{Conclusion}

In this paper, we have introduced an episode-based training paradigm to enhance the adaptability of the model for zero-shot learning. It divides the training process into a collection of episodes, each of which mimics a fake zero-shot classification task. By training multiple episodes, the model accumulates a wealth of ensemble experiences on predicting the mimetic unseen classes, which generalizes well on the real unseen classes. Under this training paradigm, we have proposed an effective generative model to align the visual-semantic consistency paired with a parameter-economic multi-modal cross-entropy loss. The comprehensive results on four benchmark datasets demonstrate that the proposed model achieves the new state-of-the-art and beats the competitors by large margins.

\noindent
\textbf{Acknowledgement.}
This work is supported in part by NSFC (61672456,U19B2043), Zhejiang Lab (2018EC0ZX01-2), the fundamental research funds for central universities in China (No. 2017FZA5007), Artificial Intelligence Research Foundation of Baidu Inc., the Key Program of Zhejiang Province, China (No. 2015C01027), the funding from HIKVision and Horizon Robotics, and ZJU Converging Media Computing Lab.


%
%

{\small
\bibliographystyle{ieee_fullname}
\bibliography{egbib}
}

\end{document}